%% file: main.tex
\documentclass{article}
\usepackage[final]{colm2026_conference}

\usepackage{microtype}
\usepackage{graphicx}
\usepackage{subcaption}
\usepackage{booktabs}
\usepackage{wrapfig}
\usepackage{booktabs}
\usepackage{hyperref}
\usepackage{url}
\usepackage{array} 
\let\cite\citep
\usepackage{lineno}

\definecolor{darkblue}{rgb}{0, 0, 0.5}
\hypersetup{colorlinks=true, citecolor=darkblue, linkcolor=darkblue, urlcolor=darkblue}

\usepackage{amsmath}
\usepackage{amssymb}
\usepackage{mathtools}
\usepackage{amsthm}

\usepackage[capitalize,noabbrev]{cleveref}

\theoremstyle{plain}

\theoremstyle{definition}

\theoremstyle{remark}

\usepackage[textsize=tiny]{todonotes}

\usepackage{color}
\usepackage{tabularx}
\usepackage{multirow}
\usepackage{tcolorbox}
\tcbuselibrary{skins}
\usepackage{float}
\usepackage{xspace}
\usepackage{bm}
\usepackage{arydshln}
\usepackage{pifont}
\newcommand{\cmark}{\ding{51}}%
\newcommand{\xmark}{\ding{55}}%
\usepackage{enumitem}

\newcommand{\ours}{Sci-VBench\xspace}

\newcommand{\ndisciplines}{four\xspace}
\newcommand{\nsubjects}{60\xspace}

\newcommand{\nmodels}{16\xspace}
\newcommand{\nproprietarymodel}{eight\xspace}
\newcommand{\nopensourcemodel}{eight\xspace}
\newcommand{\judge}{MLLM-as-Judge\xspace}
\newcommand{\nexample}{1,253\xspace}
\newcommand{\ntestmini}{150\xspace}
\newcommand{\nrefvideos}{1,500\xspace}
\newcommand{\nreratevideos}{300\xspace}
\newcommand{\nannotators}{61\xspace}
\newcommand{\eg}{\hbox{\emph{e.g.,}}\xspace}
\newcommand{\ie}{\hbox{\emph{i.e.,}}\xspace}

\newcommand{\huggingface}{\raisebox{-1.5pt}{\includegraphics[height=1.05em]{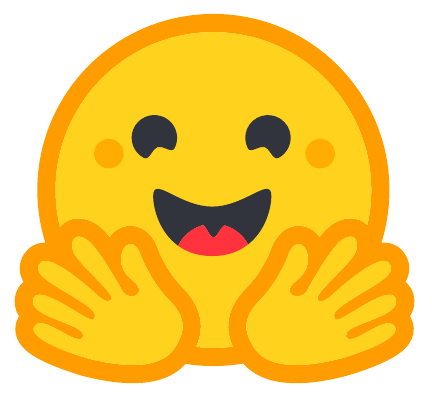}}\xspace}
\newcommand{\github}{\raisebox{-1.5pt}{\includegraphics[height=1.05em]{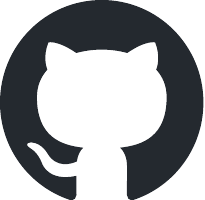}}\xspace}

\title{\ours: Evaluating Knowledge- and Reasoning-Intensive Video Generation in Science Domains}

\author{
Diandian Zhang$^{1*}$\quad
Tingyu Song$^{2*}$\quad
Lin Fu$^{1*}$\quad
Zheyuan Yang$^{3}$\quad
Yilun Zhao$^{4}$
\vspace{10pt}\\
$^1$ Zhejiang University \quad
$^2$ UCAS \quad
$^3$ Tongji University \quad
$^4$ Yale University
}

\begin{document}

\ifcolmsubmission
\linenumbers
\fi

\maketitle

{\renewcommand{\thefootnote}{}\footnotetext{$^*$Equal contributions. Correspondence to: Tingyu Song (\texttt{songtingyu23@mails.ucas.ac.cn}), Yilun Zhao (\texttt{yilun.zhao@yale.edu}).}}

\begin{abstract}
\input{main/0-abstract}
\end{abstract}

\input{figure_tex/figure1}
\clearpage

\input{main/1-intro}
\input{main/2-related_work}

\input{main/3-dataset}

\input{main/3-1-evaluation_procotol}
\input{main/5-exp}
\input{main/6-conclusion}

\begingroup
\catcode`\&=12
\bibliography{colm2026_conference, custom}
\endgroup
\bibliographystyle{colm2026_conference}

\appendix
\input{appendix/main}

\end{document}

%% file: main/0-abstract.tex
We introduce \ours, a comprehensive benchmark for evaluating knowledge- and reasoning-intensive video generation across scientific domains. It contains \nexample expert-annotated examples spanning \nsubjects subjects across \ndisciplines core disciplines: Natural Science, Healthcare, Humanities \& Social Sciences, and Engineering. Each example requires models to generate temporally rich videos that demand scientific reasoning and knowledge-grounded synthesis, going beyond surface-level visual plausibility.
We further establish a rubric-based evaluation protocol. Our analysis shows that, under this protocol, both non-expert human evaluators and \judge systems can achieve relatively high agreement with expert judgments, supporting reproducible evaluation at scale.
We benchmark \nmodels frontier proprietary and open-source models and find that, while automatic perceptual-quality scores cluster tightly across systems, performance on Prompt Grounding and Scientific and Causal Correctness varies substantially, with a pronounced proprietary--open-source gap. These findings show that advances in visual realism have not yet translated into reliable modeling of scientific and causal dynamics.

\begin{center}
\begin{tabular}{clcl}
\huggingface & \href{https://huggingface.co/Sci-VBench} {\path{Sci-VBench}} &
\github & \href{https://github.com/Sci-VBench/Sci-VBench}{\path{Sci-VBench}}\\
\end{tabular}
\end{center}

%% file: figure_tex/figure1.tex
\begin{figure}[H]
    \centering
    \includegraphics[width=\linewidth]{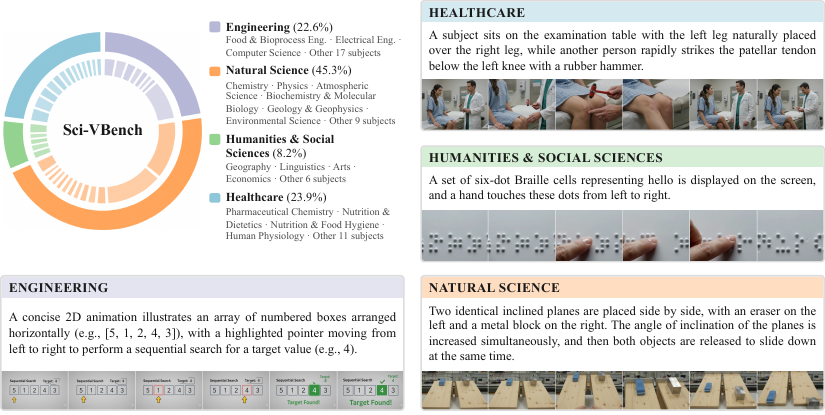}
    \caption{Overview of the \ours dataset. \textbf{Top left:} distribution of the \nexample expert-annotated examples over \nsubjects subjects across \ndisciplines core disciplines. \textbf{Remaining panels:} representative prompt--video examples from each discipline, generated by Gemini-Omni-Flash, where faithful generation requires grounding in the underlying scientific mechanism.}
\label{fig:fig1}
\end{figure}

%% file: main/1-intro.tex
\section{Introduction}
Video generative models have advanced rapidly in visual fidelity, motion coherence, and controllability~\cite{DBLP:journals/corr/abs-2507-16869, sora2, veo3, wan26}. These gains shift the central question from whether a generated video looks plausible to whether it faithfully realizes the events, constraints, and temporal dependencies specified by the prompt. This question is particularly consequential in science-domain settings, where perceptual plausibility can mask fundamental errors in the underlying process. A video may look convincing while violating a conservation law, reversing a causal relation, or depicting an impossible state transition~\cite{DBLP:conf/iclr/BansalLXZYBJSCG25, DBLP:conf/icml/MengLTLSZ0L025}. Reliable science-domain video generation therefore requires more than visual realism: it requires knowledge-grounded, temporally consistent synthesis that preserves mechanistic fidelity.

Evaluation, however, has not kept pace with this shift. Existing benchmarks provide increasingly broad coverage of perceptual quality, prompt alignment, and compositionality~\cite{DBLP:conf/cvpr/HuangHYZS0Z0JCW24, DBLP:journals/corr/abs-2503-21755, DBLP:journals/pami/HuangZXHYDMCSJWCCWLQL26, DBLP:conf/cvpr/SunHL0XLL25}, while reasoning-oriented suites largely focus on generic physical commonsense~\cite{DBLP:conf/icml/MengLTLSZ0L025, DBLP:conf/iclr/BansalLXZYBJSCG25, DBLP:journals/corr/abs-2503-06800, DBLP:journals/corr/abs-2502-20694}. Meanwhile, knowledge- and reasoning-intensive evaluation has primarily been studied for video understanding~\cite{DBLP:conf/cvpr/00010XHGLHCLXWS25, DBLP:journals/corr/abs-2501-13826, DBLP:journals/corr/abs-2510-08559, DBLP:journals/corr/abs-2606-05259, DBLP:conf/iclr/ShangguanLDZ0FC25}. Recent work has begun to examine scientific reasoning in video generation~\cite{DBLP:journals/corr/abs-2512-02942}, but a multidisciplinary benchmark that tests mechanisms across scientific domains and makes expert-defined evaluation criteria portable across raters is still missing.

To address this gap, we introduce \ours, a benchmark for knowledge- and reasoning-intensive, visually verifiable video generation across scientific domains. It comprises \nexample expert-authored and independently reviewed generation tasks spanning \nsubjects subjects across \ndisciplines core disciplines: Natural Science, Healthcare, Humanities \& Social Sciences, and Engineering. Each task pairs a minimal generation prompt with an expert-authored evaluation specification consisting of (i) a reference guide that identifies the target concept and expected phase-based phenomena and (ii) a detailed rubric with observable 1--5 scoring anchors. This design requires models to infer and render the underlying mechanism while providing evaluators with a consistent expert-defined standard.

We establish and validate a scalable rubric-based evaluation protocol. Expert human ratings serve as the reference labels. For automatic benchmarking, we combine VBench-based Vision Tools (VT) for Low-level Perceptual Fidelity with rubric-conditioned MLLM-as-judge scoring for Prompt Grounding, Scientific and Causal Correctness, and Spatiotemporal Consistency. A controlled human study shows that providing the evaluation specification substantially improves non-expert agreement with experts, while independent expert re-rating confirms the stability of the reference judgments (Cohen's $\kappa = 0.842$). Across multiple MLLM evaluators, our rubric-conditioned protocol also aligns more closely with expert judgments than prior video scoring methods~\cite{DBLP:conf/emnlp/HeJZKSSCCJAWDNL24, DBLP:journals/corr/abs-2509-22799, DBLP:conf/iccv/GuanLSZLLCS25}.

\looseness=-1 We evaluate \nmodels frontier proprietary and open-source text-to-video models on \ours. Both expert and automatic evaluations reveal a pronounced proprietary--open-source gap concentrated in Prompt Grounding and Scientific and Causal Correctness, whereas the automatic perceptual-quality proxy is nearly flat across systems. Qualitative analysis further shows that even leading models produce systematic mechanistic errors despite visually convincing outputs. Prompt rewriting improves Prompt Grounding and Scientific and Causal Correctness but has limited effect on Spatiotemporal Consistency, suggesting that many temporal failures arise from limitations of the underlying generators rather than underspecified instructions. 

%% file: main/2-related_work.tex
\section{Related Work}

\subsection{Video Generative Models}
Recent progress in video generation~\cite{DBLP:journals/corr/abs-2507-16869} has been propelled by the synergy of diffusion models and large-scale transformer architectures, carrying text-to-video (T2V) systems from the short, low-fidelity clips of early models~\cite{DBLP:conf/iclr/LiLZ0PCW25, DBLP:journals/corr/abs-2403-14773} to the high-fidelity, temporally coherent outputs of current video foundation models~\cite{sora2, veo3}.
Consequently, a significant focus of current research~\cite{DBLP:conf/cvpr/Xue00025, DBLP:conf/nips/GillmanHFALSS25, DBLP:journals/corr/abs-2505-04512} is on enhancing their adherence to physical laws and commonsense reasoning.
\looseness=-1 Adjacent work has studied scientific multimodal reasoning and static scientific visual generation~\cite{DBLP:conf/acl/WangSKC025, DBLP:conf/acl/ChenZWHPC26, DBLP:conf/acl/0001WLC25, DBLP:conf/emnlp/Wang0Z00C25}.

\subsection{Evaluation of Video Generative Models} 
\label{sec:related_work_benchmark}

Current evaluation benchmarks~\cite{DBLP:conf/cvpr/HuangHYZS0Z0JCW24, DBLP:conf/emnlp/HeJZKSSCCJAWDNL24, DBLP:journals/corr/abs-2502-20694} for video generation primarily assess foundational capabilities such as perceptual fidelity and motion quality. Some evaluation frameworks~\cite{DBLP:journals/pami/HuangZXHYDMCSJWCCWLQL26, DBLP:journals/corr/abs-2503-21755, DBLP:conf/cvpr/SunHL0XLL25} further probe compositionality, object interaction, and higher-level prompt following. These benchmarks are not designed to test whether a video remains faithful to domain-specific mechanisms and rubric-verifiable scientific outcomes.
Several recent benchmarks~\cite{DBLP:conf/iclr/BansalLXZYBJSCG25,DBLP:journals/corr/abs-2503-06800, DBLP:journals/corr/abs-2502-20694} move toward reasoning-focused evaluation by testing whether generated videos obey physical laws, but they remain grounded in generic physical commonsense rather than science-domain knowledge.
In parallel, prior work finds persistent reliability gaps in model-based evaluation for AI-generated videos and complex scientific tasks~\cite{DBLP:conf/acl/SongHG025, DBLP:conf/acl/0001CX0W0VC25, DBLP:conf/nips/ZhaoZHWBLTCDBZH25}, underscoring the need for structured evaluation specifications such as the per-example rubrics used in \ours.
The closest concurrent benchmark to ours is VideoScience-Bench~\cite{DBLP:journals/corr/abs-2512-02942}, which evaluates scientific phenomena in physics and chemistry through 200 expert-annotated prompts. In contrast, \ours spans \nsubjects subjects, provides reusable per-example reference guides and rubrics, and studies evaluation portability across expert, non-expert, and MLLM-based raters.
\autoref{tab:benchmark_comparison} summarizes these differences, comparing \ours with existing video-generation benchmarks in terms of domain coverage, evaluation focus, expert involvement, and released evaluation specifications.
\input{tables/benchmark_comparison}

%% file: tables/benchmark_comparison.tex
\begin{table*}[t]
\centering
\footnotesize
\setlength{\tabcolsep}{4pt}
\renewcommand{\arraystretch}{1.0}
\resizebox{\textwidth}{!}{%
\begin{tabular}{l l l r c c}
\toprule
\textbf{Benchmark} & \textbf{Domain} & \textbf{Focus} & \textbf{Examples} & \textbf{Expert} & \textbf{Eval Spec} \\
\midrule
VBench & General & Video quality & 946 & \xmark & \xmark \\
PhyGenBench & Physics & Physical commonsense & 160 & \xmark & \xmark \\
VideoPhy & Physics & Physical commonsense & 688 & \xmark & \xmark \\
VideoPhy-2 & Physics & Action-centric physical commonsense & 3,940 & \xmark & \xmark \\
WorldModelBench & General & World modeling & 350 & \xmark & \xmark \\
V-ReasonBench & General & Structured, spatial, pattern, physical & 326 & \xmark & \xmark \\
VideoScience-Bench & Phys./Chem. & Scientific reasoning & 200 & \cmark & \xmark \\
\midrule
\textbf{\ours} & \nsubjects sci.\ subjects & Knowledge-, Reasoning-intensive & \nexample & \cmark & \cmark \\
\bottomrule
\end{tabular}%
}
\caption{Comparing existing video-generation benchmarks with \ours. ``Expert'' indicates whether domain experts are involved in benchmark construction or annotation. ``Eval Spec'' denotes released, reusable per-example evaluation guides or rubrics.}
\label{tab:benchmark_comparison}
\end{table*}

%% file: main/3-dataset.tex
\section{\ours Benchmark}

To ensure high data quality and rigorous assessment, \ours is constructed around four core desiderata:
\textbf{(1) Breadth of domain knowledge:} \nexample examples spanning \nsubjects subjects across \ndisciplines disciplines, from astrophysics to public policy.
\textbf{(2) Depth of expert-level reasoning:} every prompt is authored by a domain expert, so generating it correctly demands grounded scientific understanding and multi-step causal reasoning.
\textbf{(3) Completeness of spatiotemporal reasoning:} the target phenomenon must unfold over time, so no single well-chosen frame can satisfy the prompt.
\textbf{(4) Reliable and reproducible evaluation:} a rubric-based protocol with per-example scoring anchors, released so that other groups can apply the same standard without recruiting domain experts.

In the following subsections, we first describe the evaluation dimensions and then detail the benchmark construction pipeline, with an overview shown in \autoref{fig:fig2}.

\subsection{Evaluation Dimensions} \label{sec:eval_dimension}
\emph{Low-level Perceptual Fidelity} and \emph{Prompt Grounding} follow established concerns in prior video-generation evaluation, while \emph{Scientific and Causal Correctness} and \emph{Spatiotemporal Consistency} isolate mechanism-level failures that generic quality or prompt-alignment criteria do not capture:
\textbf{(1) Low-level Perceptual Fidelity:} the perceptual quality of the synthesized video, including temporal coherence, motion dynamics, and the visual quality of individual frames.
\textbf{(2) Prompt Grounding:} whether the video faithfully instantiates the explicit conditions stated in the prompt, such as the presence and identity of key entities, their initial states, spatial arrangement, and any required instruments.
\textbf{(3) Scientific and Causal Correctness:} whether the generated dynamics are consistent with the domain knowledge and causal mechanisms targeted by the prompt, as specified by the per-example reference guide.
\textbf{(4) Spatiotemporal Consistency:} whether the video maintains the coherent temporal evolution and stable spatial relationships needed for the intended mechanism to be interpretable.

\input{figure_tex/figure2}
\subsection{Preliminary Setup for Benchmark Construction}
\paragraph{Subject Selection.} \label{sec:data_subject}
To ensure broad, faithful coverage of knowledge- and reasoning-intensive video generation across diverse disciplines, we conducted a user study with 133 undergraduate and graduate students to inform subject selection. Participants are asked to curate two video prompts requiring expert-level reasoning on topics related to their field of study and to provide feedback on their experiences during the curation process. The authors then manually analyzed the collected prompts together with the corresponding videos generated by Sora~\cite{sora2} and Wan 2.2~\cite{DBLP:journals/corr/abs-2503-20314}, and selected the \nsubjects subjects (listed in \autoref{tab:complete-subject-grid} in Appendix~\ref{app:subject_selection}) across \ndisciplines disciplines whose core concepts are both expert-level and verifiable from video evidence.

\paragraph{Expert Annotator Recruitment and Training.} 
Each subject is assigned to annotators with matching expertise, and every example is authored by one annotator and independently reviewed by another (\Cref{sec:quality_control}).
We include a total of \nannotators expert annotators (detailed biographies are presented in Appendix~\ref{app:annotator}); based on their current academic status, this pool comprises 11 undergraduate students, 45 graduate students, and five of the authors. All the annotators also participated in our initial user study.
Each annotator completes a training session on the annotation protocol before contributing examples.

\subsection{Video Prompt Annotation}\label{sec:video_prompt}
We annotate video prompts through a textbook-guided pipeline.
Specifically, for each subject, expert annotators first select a target concept (or tightly coupled set of concepts) from canonical textbooks and course materials that is representative of the subject's core curriculum and naturally expressed through observable spatiotemporal dynamics. 
Accordingly, \ours excludes expert concepts whose correctness cannot be verified from video evidence alone. We require that selected concepts have mechanism-governed visual realizations (\eg reaction dynamics in chemistry, conservation-driven interactions in engineering, and intervention response in healthcare), so that videos may look superficially plausible yet still produce clear and systematic deviations when the underlying principles are violated.
\looseness=-1 For each selected concept, annotators craft a minimal prompt that specifies only the observable initial setup and any explicit intervention or task objective, omitting the expected mechanistic trajectory and the key phenomena to be generated, so that a model must infer the mechanism from the setup rather than reproduce an outcome the prompt already describes.

\subsection{Evaluation Specification Curation} \label{sec:rubric_annotation}
Constructing \ours requires domain experts, but future users of the benchmark cannot be expected to recruit them at scale. We therefore release an evaluation specification with each prompt, pairing a \textbf{high-level reference guide} that fixes how the task should be interpreted with a detailed \textbf{evaluation rubric} that operationalizes scoring under that interpretation, using 1--5 anchors per dimension. This externalizes the expert knowledge evaluation requires, so that non-experts or MLLM judges can score generated videos under the same standard; we test that portability empirically in \Cref{sec:eval_reliability}.

\paragraph{High-level Reference Guide Annotation.}
We instruct annotators to write a high-level reference guide for each prompt that (i) specifies the target concept(s) and the minimal mechanistic assumptions required for the scenario to be well-defined, and (ii) summarizes the expected phenomena as a concise phase-based storyline, including key causal transitions and any visibility/viewpoint constraints needed for verification.
Moreover, to reduce terminology barriers for downstream evaluators applying the released specification, we instruct annotators to anticipate the knowledge gaps of a non-expert verifier and selectively provide brief clarifications. 

\paragraph{Evaluation Rubric Annotation.}
We instruct annotators to produce a detailed evaluation rubric aligned with our evaluation dimensions (discussed in \Cref{sec:eval_dimension}), defining 1--5 scoring anchors per dimension and tying each anchor to observable evidence. The rubric specifies what evidence is sufficient for full credit, what constitutes partial correctness, and which violations or omissions warrant low scores. Prompt Grounding, Scientific and Causal Correctness, and Spatiotemporal Consistency receive per-example rubrics, since what counts as evidence depends on the mechanism the example targets. Low-level Perceptual Fidelity instead uses a single rubric shared by every example: its anchors describe generic video-quality properties that do not depend on the scientific content.

\subsection{Data Quality Control} \label{sec:quality_control}
\input{tables/data-stat}
\autoref{tab:data_stats_lengths} reports per-discipline example counts together with the average length of each prompt, reference guide, and rubric.
Every example is reviewed in a second pass by an independent domain expert, who checks that the prompt is clear and fully specified, that the intended mechanism is visually testable in the described scene, and that the prompt, reference guide, and rubric are mutually consistent. The reviewer revises any example that fails these checks, and an author verifies the revision before the example is finalized. We further audit 200 randomly sampled examples by having a second annotator write an independent specification for the same prompt and two independent scorers grade the same video under both. Agreement is high on every reasoning dimension (quadratic weighted $\kappa$ = 0.79 for Prompt Grounding, 0.75 for Scientific and Causal Correctness, and 0.73 for Spatiotemporal Consistency), indicating that a score is determined by the released specification rather than by who wrote it. Appendix~\ref{app:quality} details the design.
\ours is released with two evaluation splits: \textbf{full}, containing all \nexample examples, and \textbf{\emph{testmini}}, a fixed subset of \ntestmini examples (37 Engineering, 16 Healthcare, 78 Natural Science, and 19 Humanities \& Social Sciences) that supports rapid iteration and cost-constrained evaluation; in our experiments, open-source models are evaluated on both splits, while proprietary systems are evaluated on \emph{testmini} only, as generating the full benchmark through commercial APIs is prohibitively expensive (\Cref{sec:evaluated_models}).

%% file: figure_tex/figure2.tex
\begin{figure*}[!t]
    \centering
    \includegraphics[width=\textwidth]{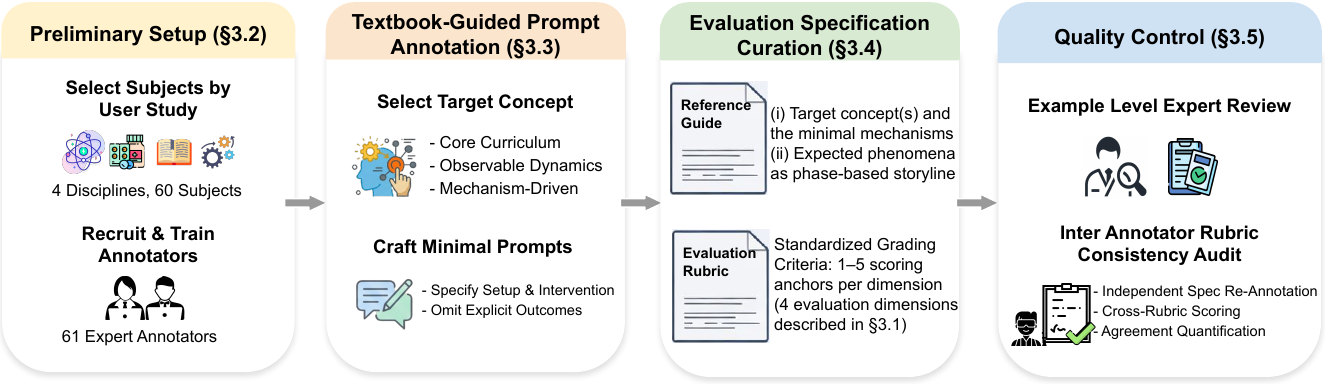}
    \caption{Overview of the \ours benchmark construction process.}
    \label{fig:fig2}
\end{figure*}

%% file: tables/data-stat.tex
\begin{wraptable}{r}{0.48\textwidth}
\centering
\small
\resizebox{0.48\textwidth}{!}{%
\setlength{\tabcolsep}{2pt} 
\renewcommand\arraystretch{1.15}

\begin{tabular}{lr*{3}{>{\raggedleft\arraybackslash}p{0.1\textwidth}}}
\toprule
\textbf{Feature} & \textbf{Eng.} & \textbf{Health.} & \textbf{Soc.} & \textbf{Nat.} \\
\midrule
\textbf{\# Example} & 283 & 300 & 103 & 567 \\
\midrule
\textbf{Length} \\
\quad Video Prompt & 42.6 & 40.5 & 35.8 & 33.3 \\
\noalign{\vskip 0.5ex}\hdashline\noalign{\vskip 0.5ex}
\quad Ref. Guide & 163.3 & 161.3 & 164.5 & 160.1 \\
\noalign{\vskip 0.5ex}\hdashline\noalign{\vskip 0.5ex}
\quad LPF Rubric & 245.0 & 245.0 & 245.0 & 245.0 \\
\quad PG Rubric & 288.7 & 300.3 & 295.3 & 280.6 \\
\quad SCC Rubric & 294.7 & 292.2 & 315.2 & 283.7 \\
\quad SC Rubric & 327.6 & 344.9 & 321.5 & 313.5 \\
\bottomrule
\end{tabular}
}
\caption{\ours statistics by discipline.}
\label{tab:data_stats_lengths}
\end{wraptable}

%% file: main/3-1-evaluation_procotol.tex
\section{\ours Evaluation Protocol}

In this section, we first describe our human and automated evaluation protocols. We further present a detailed analysis of the reliability of our automated evaluation protocols.

\subsection{Human Evaluation Protocols}
Expert human ratings serve as the primary reference labels for each evaluation dimension. Specifically, for each generated video, an expert annotator in the corresponding discipline is provided with the text prompt, the generated video, and the per-example evaluation specification, and then assigns 1--5 integer scores for all four dimensions. To ensure consistent interpretation of the rubric, annotators are trained with a brief calibration set and guided to map concrete, observable cues (\eg key state transitions, causal consistency, and verifiable evidence in the video) to the corresponding score anchors. 

\subsection{Automated Evaluation Protocols}
Expert ratings are labor-intensive and do not scale to the full benchmark or to broad model comparisons. We therefore introduce an automated protocol that scores at scale while remaining anchored to the same per-example specifications used in human evaluation. For model benchmarking, LPF is reported via VBench-based Vision Tools (VT) and the remaining three dimensions via the rubric-conditioned judge.

\paragraph{Automatic Metrics for Low-level Perceptual Fidelity.}
We assess low-level perceptual fidelity with the established automatic metrics used throughout video-generation evaluation, so that this dimension stays directly comparable to prior work.
Following VBench~\cite{DBLP:conf/cvpr/HuangHYZS0Z0JCW24}, we adopt the six metrics under its \emph{Video Quality} dimension (with definitions and implementations detailed in Appendix~\ref{app:perceptual_fidelity}). We use the official VBench evaluation protocol to compute these metrics, then normalize and average them into a single composite score, linearly mapped onto the same 1--5 range as the rubric-based dimensions, which we report as the Vision Tools (VT) score.

\paragraph{MLLM-as-Judge for All Other Dimensions.} 
We evaluate Prompt Grounding, Scientific and Causal Correctness, and Spatiotemporal Consistency using a rubric-conditioned MLLM-as-judge pipeline, instantiated with the open-source Qwen3.5-397B-A17B for reproducibility and native video input.
Specifically, for a given video–prompt pair and target dimension, \judge is provided with (1) the original text prompt, (2) the generated video, (3) the per-example high-level reference guide, and (4) the 1--5 anchored rubric for that dimension only, which keeps evidence for one dimension from bleeding into another. 
\judge is required to state a brief justification grounded in the video before assigning a single 1--5 integer score for that dimension. The single-dimension evaluation prompt is shown in \autoref{fig:judge_prompt} in Appendix~\ref{app:prompt_templates}.
To reduce run-to-run variance, every video--dimension pair is scored in three independent judge runs, and we report the mean of the three scores.

\subsection{Measuring Reliability of Evaluation Protocol}\label{sec:eval_reliability}
We assess the reliability of our evaluation protocol by measuring agreement between human experts and (1) human non-experts, with and without evaluation specifications, and (2) MLLM judges provided with evaluation specifications.

\paragraph{Collecting Reference Expert Ratings.}
Every \emph{testmini} video is rated by an expert, and Table~\ref{tab: total-results} reports those ratings for all \nmodels models. For the reliability analysis in this section we use the \nrefvideos videos generated by the ten systems released on or before January 2026. For each video, we ask an expert annotator from the same discipline, who did not author the corresponding prompt or evaluation specification, to rate video quality on each dimension using a 1--5 scale under the standardized interface described above. These scores are the reference labels for every correlation reported in \autoref{tab:eval-agree}. To assess the stability of the reference labels, we randomly sample \nreratevideos of them and ask a second independent expert from the same discipline, again not an author of the example, to re-rate them under the same protocol. Across the paired expert ratings, we obtain Cohen's $\kappa$ = 0.842, indicating strong inter-expert agreement.

\paragraph{Analyzing Non-Expert Ratings With and Without Evaluation Specifications.}
\input{tables/meta-1}
Two separate non-expert cohorts score the same expert-scored videos: in the \emph{without evaluation specification} condition evaluators see only the text prompt and the generated video, while in the \emph{with evaluation specification} condition a different cohort additionally receives the full per-example specification (\ie the high-level reference guide and scoring rubric) and is asked to follow it strictly.
As shown in \autoref{tab:eval-agree}, providing the evaluation specification increases instance-level Pearson correlation in all four dimensions. The improvement is modest for Low-level Perceptual Fidelity (84.7 vs.\ 83.2) but substantially larger for Prompt Grounding (83.3 vs.\ 74.1), Scientific and Causal Correctness (82.5 vs.\ 68.2), and Spatiotemporal Consistency (79.6 vs.\ 70.3). With the specification, non-experts agree with experts more closely than any \judge instantiation we tested, on every dimension.

\paragraph{Analyzing \judge Ratings Across Base Evaluators.}
We next assess how closely \judge scores match those of human experts, evaluating multiple \judge instantiations that swap the underlying evaluator among open-weight video understanding models of varying scale (\ie Qwen3.5-397B-A17B, Gemma-4-31B, and Qwen3.5-9B).
As shown in \autoref{tab:eval-agree}, Qwen3.5-397B-A17B attains the highest correlations overall, and agreement broadly increases with evaluator scale. For every evaluator, alignment is markedly stronger on the reasoning-centric dimensions (Prompt Grounding and Scientific and Causal Correctness) than on Low-level Perceptual Fidelity and Spatiotemporal Consistency, whose fine-grained visual and temporal artifacts remain difficult for general-purpose MLLM judges.
We further compare our automated evaluation protocol against representative prior paradigms, including VideoScore~\cite{DBLP:conf/emnlp/HeJZKSSCCJAWDNL24}, VideoScore2~\cite{DBLP:journals/corr/abs-2509-22799}, VideoReward~\cite{DBLP:conf/nips/LiuLLYLZWWXWLYW25}, and ETVA~\cite{DBLP:conf/iccv/GuanLSZLLCS25}. As \autoref{tab:eval-agree} shows, the strongest rubric-conditioned \judge correlates more closely with expert ratings than any of them on every dimension; Appendix~\ref{app:meta} reports the full setup. 

%% file: tables/meta-1.tex
\begin{wraptable}{r}{0.45\textwidth}
\vspace{-10pt}
\centering
\small
\setlength{\tabcolsep}{3pt}
\renewcommand{\arraystretch}{1.2}
\resizebox{\linewidth}{!}{%
\begin{tabular}{lccccc}
\toprule
 & \textbf{LPF} & \textbf{PG} & \textbf{SCC} & \textbf{SC} & \textbf{Avg.} \\
\midrule
\multicolumn{6}{c}{\textbf{Measuring Reliability of Our Evaluation Protocol}} \\
Human Non-expert & & & & & \\
\quad with Eval Spec. & 84.7 & 83.3 & 82.5 & 79.6 & 82.5 \\
\quad without Eval Spec. & 83.2 & 74.1 & 68.2 & 70.3 & 74.0 \\
\noalign{\vskip 0.5ex}\hdashline\noalign{\vskip 0.5ex}
Qwen3.5-397B-A17B & 53.5 & 73.9 & 71.7 & 54.5 & 63.4 \\
Gemma-4-31B & 51.2 & 68.5 & 67.2 & 54.6 & 60.4 \\
Qwen3.5-9B & 45.9 & 64.0 & 64.4 & 48.3 & 55.7 \\
\midrule
\multicolumn{6}{c}{\textbf{Comparison with Prior Auto-Eval Methods}} \\
VideoScore   & 44.8 & 58.1 & 40.6 & 46.3 & 47.5 \\
VideoScore2  & 47.9 & 60.4 & 49.8 & 48.2 & 51.6 \\
VideoReward  & 50.7 & 55.6 & 38.9 & 44.5 & 47.4 \\
ETVA         & 31.2 & 62.7 & 64.1 & 33.6 & 47.9 \\
\bottomrule
\end{tabular}}
\caption{Instance-level Pearson correlation ($\times 100$) with expert human ratings. \textbf{LPF}: Low-level Perceptual Fidelity, \textbf{PG}: Prompt Grounding, \textbf{SCC}: Scientific and Causal Correctness, \textbf{SC}: Spatiotemporal Consistency.}
\vspace{-1.5\baselineskip}
\label{tab:eval-agree}
\end{wraptable}

%% file: main/5-exp.tex
\section{Experiment}

\subsection{Evaluated Models}\label{sec:evaluated_models}
\looseness=-1 We benchmark \nmodels frontier text-to-video models on \ours, spanning \nproprietarymodel \textbf{proprietary} systems: Sora-2~\cite{sora2}, Veo-3.1-Fast and Veo-3.1~\cite{veo3}, Kling-2.6~\cite{kling26}, Wan-2.6~\cite{wan26}, Seedance-2.0~\cite{DBLP:journals/corr/abs-2604-14148}, HappyHorse-1.1~\cite{happyhorse}, and Gemini-Omni-Flash~\cite{geminiomni}, and \nopensourcemodel \textbf{open-source} models: HunyuanVideo-1.5-480P-T2V~\cite{DBLP:journals/corr/abs-2511-18870}, LTX-2.0-19B-distilled and LTX-2.3~\cite{DBLP:journals/corr/abs-2601-03233}, LongCat-Video~\cite{DBLP:journals/corr/abs-2510-22200}, Wan2.2-5B-T2V~\cite{DBLP:journals/corr/abs-2503-20314}, CogVideoX1.5-5B~\cite{DBLP:conf/iclr/YangTZ00XYHZFYZ25}, Cosmos3-Nano~\cite{cosmos3}, and MiniMax-H3~\cite{minimaxh3}. All videos are generated from the verbatim benchmark prompts under each model's default configuration, with the per-model version, resolution, frame rate, and clip duration listed in Appendix~\ref{sec: generation-model-settings}.

\subsection{Main Results}

\input{tables/table1}
\input{figure_tex/domain_heatmap_fig}
Table~\ref{tab: total-results} reports \emph{testmini} results for all \nmodels models; complete per-dimension results on the full benchmark are provided in Appendix~\ref{app:full_results} (Table~\ref{tab: full-results}). For the open-source models, which we run on both splits, per-model full-benchmark averages differ from \emph{testmini} by at most 0.07, confirming \emph{testmini} as a faithful low-cost proxy. We highlight three main findings below.
\textbf{What separates current systems is mechanism, not appearance.}
VT is nearly flat across all \nmodels models (3.79--4.12); human \emph{Low-level Perceptual Fidelity} ratings of the same construct spread wider (2.66--3.76), so perceptual quality is saturated as far as the VBench-based proxy can resolve it rather than in absolute terms. The reasoning-centric dimensions separate models far more sharply, and do so under both protocols: automatic \emph{Scientific and Causal Correctness} ranges from 1.24 to 3.34, and human SCC from 1.12 to 3.06. What distinguishes current systems on \ours is less whether videos look right than whether they get the underlying mechanism right.
\textbf{The proprietary--open-source gap is concentrated on reasoning, not on consistency.}
Gemini-Omni-Flash attains the best automatic (3.38) and human (3.18) averages, followed by HappyHorse-1.1 and Seedance-2.0; all three surpass the strongest earlier systems, Veo-3.1-Fast, Veo-3.1, and Sora-2 (2.98--3.00), and human evaluation reproduces the automatic ordering at the top of the table. Open-source systems fall behind on exactly the reasoning dimensions: MiniMax-H3 leads that group but reaches only 1.63 on SCC, half the proprietary best. On \emph{Spatiotemporal Consistency} they are not behind at all, with Wan2.2-5B attaining the highest automatic score of any model (2.79).
\textbf{No model is uniformly strong across disciplines.}
As shown in \autoref{fig:domain-results}, no system is strongest everywhere: Gemini-Omni-Flash leads three of the four disciplines (3.33 on Engineering, 3.13 on Natural Science, and 2.95 on Humanities \& Social Sciences) but Sora-2 takes Healthcare (2.97), and open-source profiles are similarly uneven. Models whose overall averages nearly coincide can therefore differ substantially in which domain mechanisms they preserve.

\subsection{Error Analysis}
{\interlinepenalty=10000
We classify the observed errors into three major categories.
\textbf{(1) Poor Adherence to Instructions:} models misinterpret core concepts and miss the fine-grained details specified in the prompt.
\textbf{(2) Inaccurate Simulation of Scientific Principles:} models prioritize visual aesthetics over physical realism, producing factually incorrect dynamics.
\textbf{(3) Deficiencies in Temporal Coherence and Visual Quality:} videos suffer from temporal inconsistencies, such as objects changing illogically over time.
Appendix~\ref{sec: error analysis} illustrates each category with frames from the evaluated models.
\par}

\subsection{Effect of Prompt Rewriting}
\input{figure_tex/prompt_enhance_fig}
\looseness=-1 Our main results use verbatim prompts (\Cref{sec:evaluated_models}); as an ablation, we ask how much of the gap more explicit prompting recovers. We rewrite each \emph{testmini} prompt with Gemini-3-Flash, instructing it to restate the scene and the requested dynamics in more visually concrete terms without changing the scenario, and regenerate with Wan2.2-5B and HunyuanVideo-1.5 under unchanged generation and evaluation settings. The gains are ordered consistently across both models (\autoref{fig:prompt_enhance}): \emph{Scientific and Causal Correctness} (SCC) improves most (+23.3\% and +51.7\%), then \emph{Prompt Grounding} (+12.4\% and +26.7\%), while \emph{Spatiotemporal Consistency} gains least (+7.5\% and +16.5\%). The gap narrows without closing: HunyuanVideo-1.5 reaches 2.20 on SCC, above the best verbatim open-source score in Table~\ref{tab: total-results} (1.63) yet far below the strongest proprietary system (3.34). Explicit wording helps where the prompt left the mechanism implicit, but it cannot supply the mechanistic fidelity the generator lacks.

%% file: tables/table1.tex
\begin{table*}[!t]
\centering
\footnotesize
\renewcommand{\arraystretch}{1.1}
\setlength{\tabcolsep}{4pt}
\resizebox{\textwidth}{!}{%
\begin{tabular}{lcrrrrrrrrrrr}
\toprule[.1em]
\multirow{2}{*}{\textbf{Models}} & \multirow{2}{*}{\textbf{Release}} & \multirow{2}{*}{\textbf{Full Avg.}} & \multicolumn{5}{c}{\textbf{Automatic Eval}} & \multicolumn{5}{c}{\textbf{Human Eval}} \\
\cmidrule(lr){4-8}  \cmidrule(lr){9-13}
 & & & \textbf{VT} & \textbf{SC} & \textbf{PG} & \textbf{SCC} & \textbf{Avg.} & \textbf{LPF} & \textbf{SC} & \textbf{PG} & \textbf{SCC} & \textbf{Avg.} \\
\midrule
\multicolumn{13}{c}{\emph{\textbf{Proprietary Models}}} \\
\midrule
Gemini-Omni-Flash~\cite{geminiomni} & 2026-06 & -- & \textbf{4.12} & 2.53 & \textbf{3.52} & \textbf{3.34} & \textbf{3.38} & \textbf{3.76} & 2.53 & \textbf{3.37} & \textbf{3.06} & \textbf{3.18} \\
HappyHorse-1.1~\cite{happyhorse} & 2026-06 & -- & 3.96 & 2.66 & \underline{3.21} & 2.76 & \underline{3.15} & \underline{3.75} & 2.85 & \underline{3.09} & 2.57 & \underline{3.06} \\
Seedance-2.0~\cite{DBLP:journals/corr/abs-2604-14148} & 2026-02 & -- & 3.86 & 2.46 & 3.17 & \underline{3.06} & 3.14 & 3.73 & 2.32 & 3.01 & \underline{2.68} & 2.94 \\
Veo-3.1~\cite{veo3} & 2025-10 & -- & \underline{4.10} & 2.26 & 2.83 & 2.75 & 2.98 & 3.53 & 2.66 & 2.65 & 2.37 & 2.80 \\
Veo-3.1-Fast~\cite{veo3} & 2025-10 & -- & 4.03 & 2.34 & 2.92 & 2.70 & 3.00 & 3.45 & 2.71 & 2.67 & 2.34 & 2.79 \\
Sora-2~\cite{sora2} & 2025-09 & -- & 3.82 & 2.29 & 3.12 & 2.72 & 2.98 & 3.40 & 2.29 & 2.77 & 2.24 & 2.68 \\
Kling-2.6~\cite{kling26} & 2025-12 & -- & 3.92 & 2.59 & 2.60 & 1.71 & 2.70 & 3.49 & \textbf{2.91} & 2.41 & 1.65 & 2.62 \\
Wan-2.6~\cite{wan26} & 2025-12 & -- & 3.89 & 2.61 & 2.43 & 1.79 & 2.68 & 3.42 & 2.82 & 2.27 & 1.58 & 2.52 \\
\midrule
\multicolumn{13}{c}{\emph{\textbf{Open-source Models}}} \\
\midrule
MiniMax-H3~\cite{minimaxh3} & 2026-08 & \textbf{2.61} & 3.94 & 2.62 & 2.52 & 1.63 & 2.68 & 3.64 & 2.83 & 2.29 & 1.54 & 2.58 \\
HunyuanVideo-1.5~\cite{DBLP:journals/corr/abs-2511-18870} & 2025-11 & \underline{2.43} & 3.79 & 2.66 & 1.87 & 1.45 & 2.44 & 3.45 & \underline{2.90} & 1.80 & 1.21 & 2.34 \\
LongCat-Video~\cite{DBLP:journals/corr/abs-2510-22200} & 2025-10 & 2.34 & 3.92 & 2.46 & 1.76 & 1.24 & 2.34 & 3.45 & 2.76 & 1.75 & 1.13 & 2.27 \\
Cosmos3-Nano~\cite{cosmos3} & 2026-06 & \underline{2.43} & 4.05 & \underline{2.73} & 1.76 & 1.27 & 2.45 & 3.37 & 2.73 & 1.67 & 1.14 & 2.23 \\
Wan2.2-5B~\cite{DBLP:journals/corr/abs-2503-20314} & 2025-07 & \underline{2.43} & 3.87 & \textbf{2.79} & 1.85 & 1.33 & 2.46 & 3.05 & 2.82 & 1.79 & 1.21 & 2.22 \\
LTX-2~\cite{DBLP:journals/corr/abs-2601-03233} & 2026-01 & 2.31 & 4.07 & 2.36 & 1.51 & 1.26 & 2.30 & 3.13 & 2.74 & 1.48 & 1.12 & 2.12 \\
CogVideoX1.5-5B~\cite{DBLP:conf/iclr/YangTZ00XYHZFYZ25} & 2024-11 & 2.30 & 3.98 & 2.11 & 1.92 & 1.40 & 2.35 & 2.68 & 2.11 & 1.75 & 1.19 & 1.93 \\
LTX-2.3~\cite{DBLP:journals/corr/abs-2601-03233} & 2026-03 & 2.27 & 3.96 & 2.16 & 1.80 & 1.28 & 2.30 & 2.66 & 2.15 & 1.70 & 1.18 & 1.92 \\
\bottomrule[.1em]
\end{tabular}
}
\caption{Model performance on the \ours \emph{testmini} split (150 examples), with \textbf{Full Avg.} reporting the automatic average on the full benchmark for open-source models. \textbf{VT} = VBench-based Vision Tools, \textbf{SC} = Spatiotemporal Consistency, \textbf{PG} = Prompt Grounding, \textbf{SCC} = Scientific and Causal Correctness, and \textbf{LPF} = Low-level Perceptual Fidelity. Both \textbf{Avg.} columns are unweighted means over the four dimensions of their block; VT is the automatic proxy for LPF. Provider filters rejected 6 Gemini-Omni-Flash prompts and 2 Seedance-2.0 prompts; affected averages use successful generations. Rows are sorted by Human Eval \textbf{Avg.} within each model group; \textbf{bold} and \underline{underline} mark the best and second-best results.}
\label{tab: total-results}
\end{table*}

%% file: figure_tex/domain_heatmap_fig.tex
\begin{wrapfigure}{r}{0.44\textwidth}
  \vspace{-1.5\baselineskip}
  \centering
  \includegraphics[width=0.43\textwidth]{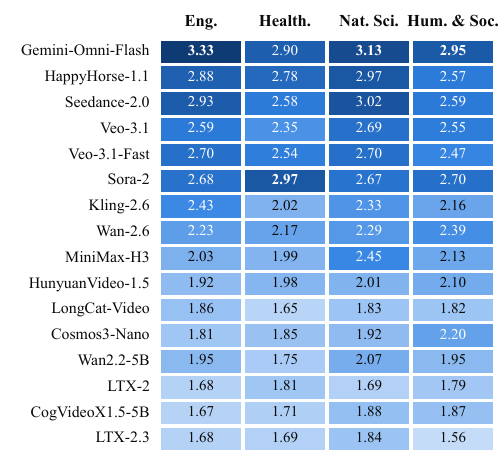}
  \vspace{-0.5\baselineskip}
  \caption{Per-discipline mean of the SC/PG/SCC judge scores on \emph{testmini}.}
  \label{fig:domain-results}
\end{wrapfigure}

%% file: figure_tex/prompt_enhance_fig.tex
\begin{wrapfigure}{r}{0.42\textwidth}
  \centering
  \vspace{-\baselineskip}
  \includegraphics[width=0.4\textwidth]{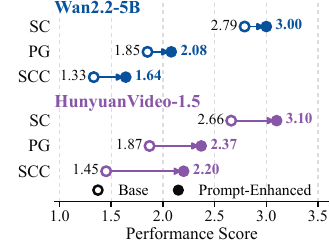}
  \caption{Performance enhancement via prompt rewriting on \emph{testmini}.}
  \label{fig:prompt_enhance}
\end{wrapfigure}

%% file: main/6-conclusion.tex
\section{Conclusion}
\ours fills a key gap in evaluating text-to-video models as expert-domain ``world simulators'' by providing \nexample expert-authored prompts spanning \nsubjects subjects across \ndisciplines disciplines, where success hinges on mechanistic reasoning and temporally coherent causal dynamics rather than surface-level realism. Benchmarking \nmodels frontier proprietary and open-source models on \ours reveals a persistent gap between perceptual realism and scientific validity, together with a substantial proprietary/open-source performance gap on the reasoning-centric dimensions. The observed failure modes, in turn, point to concrete opportunities to improve instruction adherence, mechanistic consistency, and spatiotemporal stability in expert-domain generation.

%% file: appendix/main.tex
\input{appendix/A}

\input{appendix/experiment_results}

%% file: appendix/A.tex
\clearpage
\section{\ours Dataset}
\subsection{Subject Selection} \label{app:subject_selection}
\input{tables/subject_list_table}

\subsection{Data Quality Control}\label{app:quality}

To test whether independently authored evaluation specifications support consistent judgments across independent evaluators, we conduct a targeted audit on a random sample of 200 examples. For each example, we recruit a second annotator from the same subject area to construct the full evaluation specification independently, including the high-level reference guide and the 1--5 anchored rubrics for all dimensions, without access to the first annotator's specification. This produces two independently authored specifications for the same prompt.
To quantify consistency, for each sampled example we randomly select one generated video from a randomly chosen model and recruit two independent scorers from the same subject area, neither of whom authored either specification. Each scorer evaluates the same video under both independently authored specifications in counterbalanced order, yielding a crossed scorer--rubric design. We then compute dimension-wise agreement across the resulting paired score sets using quadratic weighted Cohen's kappa: Prompt Grounding ($\kappa$ = 0.79), Scientific and Causal Correctness ($\kappa$ = 0.75), and Spatiotemporal Consistency ($\kappa$ = 0.73). These results indicate that independently authored specifications yield strongly consistent judgments. 

\clearpage
\subsection{Video Generation Models}
\label{sec: generation-model-settings}
We provide the detailed settings of the video generation models in ~\autoref{tab: video-model-specifications}. 
\input{tables/model_settings}

\clearpage
\subsection{Annotator Information}
\label{app:annotator}
We list annotator biographies in \autoref{tab:annotators-part1} and \autoref{tab:annotators-part2}; entries for annotators who are also authors are withheld to preserve anonymity.
\input{tables/annotator}
\input{tables/annotator-2}

\clearpage
\section{\ours Evaluation Protocol}
\subsection{Low-level Video Quality: Definitions and Implementation Details (Adapted from VBench)}
\label{app:perceptual_fidelity}
\input{appendix/Video_Quality/video_quality}

\subsection{Comparing with Prior Video Scoring Methods}\label{app:meta}
We adopt four methods:
\textbf{(1) VideoScore}~\cite{DBLP:conf/emnlp/HeJZKSSCCJAWDNL24} trains a learned automatic evaluator on VideoFeedback by fine-tuning a video-capable VLM to regress five aspect scores: Visual Quality, Temporal Consistency, Dynamic Degree, Text-to-Video Alignment, and Factual Consistency. We run the released evaluator on each generated video to obtain the five aspect scores, then compute their unweighted average as the overall VideoScore.
\textbf{(2) VideoScore2}~\cite{DBLP:journals/corr/abs-2509-22799} is a ``think-before-scoring'' judge trained on VideoFeedback2 with human scores plus reasoning traces, using a two-stage SFT+RL pipeline, and it outputs three scores: visual quality, text alignment, and physical/common-sense consistency. We apply the released model to each video and take the unweighted average of the three reported dimension scores as the overall VideoScore2.
\textbf{(3) VideoReward}~\cite{DBLP:conf/nips/LiuLLYLZWWXWLYW25} is a VLM-based reward model trained from a large-scale human preference dataset over three dimensions: Visual Quality (VQ), Motion Quality (MQ), and Text Alignment (TA). We compute per-video VQ/MQ/TA with VideoReward and use their unweighted average as the overall VideoReward score.
\textbf{(4) ETVA}~\cite{DBLP:conf/iccv/GuanLSZLLCS25} parses video prompts into semantic scene graphs, generating fine-grained atomic questions, and scoring videos via knowledge-augmented, multi-stage question answering. We follow the official protocol to compute ETVA's per-video alignment score (aggregated over atomic questions) and use it as the ETVA score.

As shown in \Cref{tab:eval-agree}, our rubric-conditioned \judge
aligns more closely with expert ratings than the compared prior auto-eval methods on the reasoning-centric dimensions, achieving the highest instance-level Pearson correlation among the compared automatic methods on Prompt Grounding, Scientific and Causal Correctness, and Spatiotemporal Consistency. Among prior methods, ETVA is comparatively competitive on Prompt Grounding and Scientific and Causal Correctness but degrades sharply on Low-level Perceptual Fidelity and Spatiotemporal Consistency.

\clearpage
\input{appendix/C}

%% file: tables/subject_list_table.tex
\begin{table*}[h]
\centering

\setlength{\tabcolsep}{6pt}
\renewcommand\arraystretch{1.05}
\scriptsize
\resizebox{\textwidth}{!}{%
\definecolor{avggray}{gray}{0.93}
\newcolumntype{C}[1]{>{\arraybackslash}p{#1}}
\newcolumntype{G}[1]{>{\centering\arraybackslash\columncolor{avggray}}p{#1}}
\providecommand{\up}[1]{\textsuperscript{\scriptsize\,#1}}
\begin{tabular}{C{0.24\textwidth} C{0.24\textwidth} C{0.24\textwidth} C{0.24\textwidth}}
\toprule[1pt]
\textbf{Natural Sciences (15)} & \textbf{Engineering (20)} & \textbf{Healthcare (15)} & \textbf{Humanities \& Social Sciences (10)} \\
\midrule[0.7pt]

Astrophysics \& Astronomy & Aerospace Engineering & Anthropotomy & Archaeology \\
Atmospheric Science & Agricultural Engineering & Basic Immunology & Architecture \& Urban Planning \\
Biochemistry \& Molecular Biology & Biomedical Engineering & Clinical Laboratory Science & Arts \\
Biophysics & Chemical Engineering & Digestive Physiology & Business Administration \\
Cell \& Developmental Biology & Civil \& Environmental Engineering & Human Physiology & Economics \\
Chemistry & Communications Engineering & Imaging Medicine & Education \\
Earth Science & Computer Science & Medicine & Geography \\
Ecology \& Evolutionary Biology & Electrical Engineering & Microbiology-Immunology & Linguistics \\
Environmental Science & Energy Engineering & Nutrition \& Dietetics & Public Policy Administration \\
Geology \& Geophysics & Food and Bioprocess Engineering & Nutrition and Food Hygiene & Sport \\
Marine Science & Geotechnical Engineering & Pathology &  \\
Microbiology & Hydraulic Engineering & Pathophysiology &  \\
Morphology & Instrument and Meter Engineering & Pharmaceutical Chemistry &  \\
Physics & Materials Science \& Engineering & Pharmacy &  \\
Statistics \& Data Science & Mechanical Engineering & Skin Physiology &  \\
 & Network Engineering &  &  \\
 & Nuclear Engineering &  &  \\
 & Optical Engineering &  &  \\
 & Scientific Computing &  &  \\
 & Structural Engineering &  &  \\

\bottomrule[1pt]
\end{tabular}
}
\caption{Complete subject list by major disciplines. Columns list subfields under each of the four major disciplines.}
\label{tab:complete-subject-grid}
\end{table*}

%% file: tables/model_settings.tex
\begin{table*}[htbp]
\centering
\footnotesize
\setlength{\tabcolsep}{4pt}
\renewcommand\arraystretch{1.15}
\resizebox{\textwidth}{!}{%
\begin{tabular}{llllcrrrr}
\toprule
\textbf{Organization} & \textbf{Model} & \textbf{Release} & \textbf{Version} & \textbf{Access} & \textbf{Width} & \textbf{Height} & \textbf{FPS} & \textbf{Duration} \\
\midrule
\multicolumn{9}{c}{\emph{\textbf{Proprietary Models}}} \\
\midrule
OpenAI
  & Sora-2             & 2025-09 & \texttt{sora-2}                         & API & 1280 & 720  & 30 & 15s \\
\noalign{\vskip 0.3ex}\hdashline\noalign{\vskip 0.3ex}
\multirow{3}{*}{Google}
  & Veo-3.1            & 2025-10 & \texttt{veo-3.1-generate-preview}       & API & 1280 & 720  & 24 & 8s  \\
  & Veo-3.1-Fast       & 2025-10 & \texttt{veo-3.1-fast-generate-preview}  & API & 1280 & 720  & 24 & 8s  \\
  & Gemini-Omni-Flash  & 2026-06 & \texttt{gemini-omni-flash-preview}      & API & 1280 & 720  & 24 & 10s \\
\noalign{\vskip 0.3ex}\hdashline\noalign{\vskip 0.3ex}
Kuaishou
  & Kling-2.6          & 2025-12 & \texttt{kling-v2-6}                     & API & 1440 & 1440 & 24 & 10s \\
\noalign{\vskip 0.3ex}\hdashline\noalign{\vskip 0.3ex}
\multirow{2}{*}{Alibaba}
  & Wan-2.6            & 2025-12 & \texttt{wan2.6-t2v}                     & API & 1280 & 720  & 30 & 15s \\
  & HappyHorse-1.1     & 2026-06 & \texttt{happyhorse-1.1-t2v}             & API & 1280 & 720  & 24 & 15s \\
\noalign{\vskip 0.3ex}\hdashline\noalign{\vskip 0.3ex}
ByteDance Seed
  & Seedance-2.0       & 2026-02 & \texttt{doubao-seedance-2-0-260128}     & API & 1280 & 720  & 24 & 15s \\
\midrule
\multicolumn{9}{c}{\emph{\textbf{Open-source Models}}} \\
\midrule
Tencent
  & HunyuanVideo-1.5   & 2025-11 & \texttt{tencent/HunyuanVideo-1.5}       & Open & 848  & 480  & 10 & 15s \\
\noalign{\vskip 0.3ex}\hdashline\noalign{\vskip 0.3ex}
\multirow{2}{*}{Lightricks}
  & LTX-2              & 2026-01 & \texttt{Lightricks/LTX-2} (19B-distilled) & Open & 1536 & 1024 & 24 & 15s \\
  & LTX-2.3            & 2026-03 & \texttt{Lightricks/LTX-2.3}             & Open & 1536 & 1024 & 24 & 15s \\
\noalign{\vskip 0.3ex}\hdashline\noalign{\vskip 0.3ex}
Meituan
  & LongCat-Video      & 2025-10 & \texttt{meituan-longcat/LongCat-Video}  & Open & 832  & 480  & 10 & 15s \\
\noalign{\vskip 0.3ex}\hdashline\noalign{\vskip 0.3ex}
Alibaba
  & Wan2.2-5B          & 2025-07 & \texttt{Wan-AI/Wan2.2-TI2V-5B}          & Open & 1280 & 704  & 16 & 15s \\
\noalign{\vskip 0.3ex}\hdashline\noalign{\vskip 0.3ex}
Zhipu AI
  & CogVideoX1.5-5B    & 2024-11 & \texttt{THUDM/CogVideoX1.5-5B}          & Open & 1360 & 768  & 16 & 5s  \\
\noalign{\vskip 0.3ex}\hdashline\noalign{\vskip 0.3ex}
NVIDIA
  & Cosmos3-Nano       & 2026-06 & \texttt{nvidia/Cosmos3-Nano}            & Open & 1280 & 720  & 24 & 15s \\
\noalign{\vskip 0.3ex}\hdashline\noalign{\vskip 0.3ex}
MiniMax
  & MiniMax-H3         & 2026-08 & \texttt{MiniMaxAI/MiniMax-H3}           & Open & 1344 & 768  & 24 & 15s \\
\bottomrule
\end{tabular}
}
\caption{Details of the evaluated video generation models. All videos are generated from the verbatim benchmark prompts under each model's default configuration; \textbf{Duration} is the supported setting closest to our 15-second target (Veo-3.1 caps at 8s, Gemini-Omni-Flash at 10s, Kling-2.6 at 10s, and CogVideoX1.5-5B at 5s). \textbf{Version} gives the exact API model identifier for proprietary systems and the HuggingFace repository for open-source models; resolution and frame rate are measured from the generated videos. HappyHorse-1.1 clips carry a provider watermark (present in our generations even with the documented \texttt{watermark: false} request parameter).}
\label{tab: video-model-specifications}
\end{table*}

%% file: tables/annotator.tex
\begin{table*}[h]
\centering

\renewcommand{\arraystretch}{1.05}
\setlength{\tabcolsep}{4pt}
\resizebox{\textwidth}{!}{%
\begin{tabular}{c l l l c c}
\toprule
\textbf{ID} & \textbf{Year} & \textbf{Major} & \textbf{Assigned Subject(s)} & \textbf{Author?} & \textbf{Validator?} \\
\midrule

1  & 4th yr Undergraduate & Physics & Physics & \xmark & \xmark \\
2  & 4th yr Undergraduate & Chemistry & Chemistry & \xmark & \xmark \\
3  & 1st yr Master & Mathematics & Statistics \& Data Science & \xmark & \xmark \\
4  & 1st yr Master & Earth Science & Earth Science & \xmark & \xmark \\

5  & - & - & - & \cmark & \cmark \\

6  & 1st yr Master & Biology & Cell \& Developmental Biology & \xmark & \xmark \\
7  & 1st yr Master & Statistics & Statistics \& Data Science & \xmark & \xmark \\
8  & 1st yr Master & Environmental Science & Environmental Science & \xmark & \xmark \\
9  & 1st yr Master & Geology & Geology \& Geophysics & \xmark & \xmark \\

10 & 1st yr Master & Physics & Astrophysics \& Astronomy & \xmark & \cmark \\
11 & 1st yr Master & Chemistry & Biochemistry \& Molecular Biology & \xmark & \cmark \\
12 & 1st yr Master & Mathematics & Statistics \& Data Science & \xmark & \cmark \\
13 & 2nd yr Master & Biology & Ecology \& Evolutionary Biology & \xmark & \cmark \\
14 & 1st yr Master & Environmental Science & Atmospheric Science & \xmark & \cmark \\

15 & 1st yr Master & Physics & Physics & \xmark & \cmark \\
16 & 1st yr Master & Chemistry & Chemistry & \xmark & \cmark \\
17 & 1st yr Master & Mathematics & Statistics \& Data Science & \xmark & \cmark \\

18 & - & - & - & \cmark & \cmark \\

19 & 2nd yr PhD & Geophysics & Geology \& Geophysics & \xmark & \cmark \\
20 & 1st yr PhD & Neuroscience & Human Physiology & \xmark & \cmark \\
21 & 2nd yr PhD & Genetics & Biochemistry \& Molecular Biology & \xmark & \cmark \\
22 & 3rd yr PhD & Atmospheric Science & Atmospheric Science & \xmark & \cmark \\
23 & 2nd yr PhD & Paleontology & Ecology \& Evolutionary Biology & \xmark & \cmark \\

24 & 4th yr Undergraduate & Computer Science & Computer Science & \xmark & \xmark \\
25 & 1st yr Master & Electrical Engineering & Electrical Engineering & \xmark & \xmark \\
26 & 4th yr Undergraduate & Mechanical Engineering & Mechanical Engineering & \xmark & \xmark \\
27 & 1st yr Master & Chemical Engineering & Chemical Engineering & \xmark & \xmark \\
28 & 1st yr Master & Software Engineering & Computer Science & \xmark & \cmark \\
29 & 2nd yr Master & Materials Science & Materials Science \& Engineering & \xmark & \cmark \\
30 & 1st yr Master & Biomedical Engineering & Biomedical Engineering & \xmark & \cmark \\
31 & 2nd yr Master & Industrial Engineering & Mechanical Engineering & \xmark & \cmark \\

\bottomrule
\end{tabular}
}
\caption{Biographies of 61 annotators involved in \ours construction (Author biographies are hidden to protect identity confidentiality).}
\label{tab:annotators-part1}
\end{table*}

%% file: tables/annotator-2.tex
\begin{table*}[h]
\centering

\renewcommand{\arraystretch}{1.05}
\setlength{\tabcolsep}{4pt}
\resizebox{\textwidth}{!}{%
\begin{tabular}{c l l l c c}
\toprule
\textbf{ID} & \textbf{Year} & \textbf{Major} & \textbf{Assigned Subject(s)} & \textbf{Author?} & \textbf{Validator?} \\
\midrule

32 & 1st yr PhD & Electrical Engineering & Electrical Engineering & \xmark & \cmark \\

33 & - & - & - & \cmark & \cmark \\

34 & 2nd yr PhD & Computer Engineering & Scientific Computing & \xmark & \cmark \\
35 & 3rd yr PhD & Mechanical Engineering & Mechanical Engineering & \xmark & \cmark \\
36 & 2nd yr PhD & Chemical Engineering & Chemical Engineering & \xmark & \cmark \\

37 & 4th yr Undergraduate & Economics & Economics & \xmark & \xmark \\
38 & 1st yr Master & Sociology & Public Policy Administration & \xmark & \xmark \\
39 & 4th yr Undergraduate & Psychology & Education & \xmark & \xmark \\
40 & 4th yr Undergraduate & Geography & Geography & \xmark & \xmark \\
41 & 1st yr Master & Economics & Economics & \xmark & \cmark \\
42 & 1st yr Master & Political Science & Public Policy Administration & \xmark & \cmark \\
43 & 1st yr Master & Anthropology & Archaeology & \xmark & \cmark \\
44 & 2nd yr Master & Education & Education & \xmark & \cmark \\
45 & 1st yr Master & Economics & Economics & \xmark & \cmark \\
46 & 1st yr Master & Psychology & Education & \xmark & \cmark \\

47 & - & - & - & \cmark & \cmark \\

48 & 3rd yr PhD & Urban Studies & Architecture \& Urban Planning & \xmark & \cmark \\
49 & 2nd yr PhD & Linguistics & Linguistics & \xmark & \cmark \\

50 & 4th yr Undergraduate & Nursing & Medicine & \xmark & \xmark \\
51 & 4th yr Undergraduate & Public Health & Human Physiology & \xmark & \xmark \\
52 & 4th yr Undergraduate & Pharmacy & Pharmacy & \xmark & \xmark \\
53 & 4th yr Undergraduate & Nutrition & Nutrition \& Dietetics & \xmark & \xmark \\
54 & 1st yr Master & Clinical Medicine & Medicine & \xmark & \cmark \\
55 & 1st yr Master & Pharmacy & Pharmaceutical Chemistry & \xmark & \cmark \\
56 & 1st yr Master & Nutrition Science & Nutrition and Food Hygiene & \xmark & \cmark \\
57 & 2nd yr Master & Epidemiology & Pathology & \xmark & \cmark \\
58 & 1st yr Master & Public Health & Digestive Physiology & \xmark & \cmark \\

59 & - & - & - & \cmark & \cmark \\

60 & 3rd yr PhD & Biomedical Science & Microbiology-Immunology & \xmark & \cmark \\
61 & 2nd yr PhD & Health Informatics & Imaging Medicine & \xmark & \cmark \\

\bottomrule
\end{tabular}
}
\caption{Biographies of 61 annotators involved in \ours construction (Author biographies are hidden to protect identity confidentiality).}
\label{tab:annotators-part2}
\end{table*}

%% file: appendix/Video_Quality/video_quality.tex
\textbf{Subject Consistency}: Measures whether the main subject(s) in the video remain stable and coherent across frames. It evaluates whether the subject’s identity, shape, appearance, and key attributes (e.g., color, size, structure) are preserved throughout the video, without unexpected changes, distortions, or disappearance during temporal progression.

\textbf{Background Consistency}: Assesses whether the scene background remains temporally stable across frames. It focuses on the continuity of environmental elements, such as layout, lighting, and spatial structure, ensuring that the background does not flicker, shift unnaturally, or change inconsistently when no scene transition is intended.

\textbf{Motion Smoothness}: Motion Smoothness evaluates the temporal continuity and physical plausibility of motion in the video. Measure whether object movements, camera motion, and transitions between frames are smooth, continuous, and free from jitter, abrupt jumps, or unnatural temporal artifacts.

\textbf{Dynamic Degree}: Dynamic Degree reflects the intensity and richness of motion present in the video.
 Assesses whether the video contains an appropriate level of dynamic variation, such as movement, deformation, or interaction of an object, rather than being overly static or motionless. This metric does not judge correctness, but rather the amount of motion activity.

\textbf{Aesthetic Quality}: Aesthetic Quality evaluates the overall visual appeal and artistic quality of the video. It considers factors such as composition, color harmony, lighting, visual balance, and stylistic coherence, measuring how pleasing and well-structured the video appears from a human perceptual perspective.

\textbf{Imaging Quality}: Imaging Quality measures the low-level visual fidelity of the video frames. It focuses on technical aspects including sharpness, resolution, noise level, compression artifacts, blur, and rendering clarity, reflecting how clean and realistic the generated images appear at the pixel level.

%% file: appendix/C.tex
\subsection{Evaluation Prompt Templates}
\label{app:prompt_templates}
\autoref{fig:judge_prompt} shows the template that conditions \judge on one
evaluation dimension. \autoref{fig: example-prompt_rubic} shows the evaluation
specification it is conditioned on, as released with the benchmark: the verbatim
generation prompt, followed by the 1--5 anchored rubric for each judged dimension.

\input{figure_tex/mllm_as_judge_example}

\input{figure_tex/rubic_example}

%% file: figure_tex/mllm_as_judge_example.tex
\begin{figure*}[htbp]
\centering
\begin{tcolorbox}[
    colback=black!5,
    colframe=black!75,
    fonttitle=\bfseries,
    title=MLLM As Judge,
    width=\textwidth,
    arc=2mm,
    boxrule=0.5pt,
    enhanced,
    left=6pt,
    right=6pt,
    top=4pt,
    bottom=15pt,
    attach boxed title to top center={yshift=-2mm},
    boxed title style={
        size=normal,
        colback=white,
        colframe=black!75,
        coltext=black,
        arc=1mm,
    },
    coltitle=black,
    titlerule=0.5pt,
    title style={top color=white, bottom color=white}
]
\footnotesize
\setlength{\baselineskip}{1.1\baselineskip}

\noindent You are a professional video quality assessment expert. You are given a video generated by a Text-to-Video model. Your task is to score this video on 1 dimension(s) based on the provided rubric. \\
\vspace{0.35em}
\noindent \#\# Generation Prompt \\
\noindent \textless{}generation prompt, verbatim from the benchmark\textgreater{} \\
\vspace{0.35em}
\noindent \#\# Reference Guide \\
\noindent \textless{}per-example high-level reference guide: target concept, minimum mechanism, and the expected phase-based storyline\textgreater{} \\
\vspace{0.35em}
\noindent \#\# Scoring Rubric \\
\vspace{0.35em}
\noindent \#\#\# 1. \textless{}target dimension\textgreater{} \\
\noindent \textless{}dimension definition\textgreater{} \\
\noindent - Score 1: \textless{}anchor for score 1\textgreater{} \\
\noindent - Score 2: \textless{}anchor for score 2\textgreater{} \\
\noindent - Score 3: \textless{}anchor for score 3\textgreater{} \\
\noindent - Score 4: \textless{}anchor for score 4\textgreater{} \\
\noindent - Score 5: \textless{}anchor for score 5\textgreater{} \\
\vspace{0.35em}
\noindent \#\# Instructions \\
\noindent Watch the video carefully, then score the video on each of the 1 dimension(s). For each dimension, provide: \\
\noindent 1. A brief justification (2-3 sentences) \\
\noindent 2. A score from 1 to 5 \\
\vspace{0.35em}
\noindent Output your response in the following JSON format (use double quotes): \\
\noindent \{ \\
\noindent   "\textless{}target dimension\textgreater{}": \{"justification": "...", "score": N\} \\
\noindent \} \\
\vspace{0.35em}

\end{tcolorbox}
\caption{Prompt template for rubric-conditioned MLLM-as-judge scoring on a single evaluation dimension, reproduced from the scoring code. Angle-bracketed fields are filled per example; the rubric block carries only the anchors for the dimension being scored.}
\label{fig:judge_prompt}
\end{figure*}

%% file: figure_tex/rubic_example.tex
\begin{figure*}[htbp]
\centering
\begin{tcolorbox}[
    colback=black!5,
    colframe=black!75,
    fonttitle=\bfseries,
    title=Rubric Example,
    width=\textwidth,
    arc=2mm,
    boxrule=0.5pt,
    enhanced,
    left=6pt,
    right=6pt,
    top=4pt,
    bottom=10pt,
    attach boxed title to top center={yshift=-2mm},
    boxed title style={
        size=normal,
        colback=white,
        colframe=black!75,
        coltext=black,
        arc=1mm,
    },
    coltitle=black,
    titlerule=0.5pt,
    title style={top color=white, bottom color=white}
]
\scriptsize
\setlength{\baselineskip}{1.1\baselineskip}

We present a rubric example to demonstrate rubric-based scoring in \ours. The prompt below is the verbatim benchmark prompt; the anchors are the expert-authored evaluation specification released with it.\\
\vspace{0.4em}
\textbf{Prompt:} A subject sits on the examination table with the left leg naturally placed over the right leg, while another person rapidly strikes the patellar tendon below the left knee with a rubber hammer.\\
\vspace{0.3em}

pg\_specification: \{ \\
    1 point: ``No seated subject appears; the left-over-right leg crossing is absent; no reflex hammer is visible anywhere in frame; the region just below the left kneecap is never shown or is fully covered by clothing or hands.'' \\
    2 points: ``A seated subject is shown, but the legs are not crossed left-over-right (e.g., legs parallel, right-over-left, or uncrossed); or a hammer appears but is never brought near the left knee; or the region just below the left kneecap is not visible when the hammer approaches.'' \\
    3 points: ``Subject sits with the left leg crossed over the right leg and the region just below the left kneecap exposed; a rubber-headed reflex hammer is held and brought down to strike that exact region (not the kneecap itself, not the thigh muscle, not the shin).'' \\
    4 points: ``All elements of 3-score are present AND the subject's crossed-leg posture and upper body stay motionless in the moments before impact (no shifting, flinching, or leg repositioning as the hammer approaches); the left knee stays extended or nearly so, not bent more than about 45\textdegree{}; the hammer's head meets the tendon region square-on rather than at a shallow glancing angle.'' \\
    5 points: ``All elements of 4-score are present AND the seating, leg-crossing, hammer approach, and moment of contact play out as one continuous, uninterrupted shot with the left knee, hammer, and lower leg all kept in frame and unobstructed by hands, clothing, or camera angle from start through the strike; whether the leg subsequently moves is not scored at this level.'' \\
\} \\
\vspace{0.2em}

scc\_specification: \{ \\
    1 point: ``No leg movement follows the strike (the leg stays completely still); or the leg movement shown is physiologically wrong for this reflex, e.g., the hip flexes, the ankle flexes or dorsiflexes, or the leg withdraws rather than extends.'' \\
    2 points: ``The leg moves after the strike, but the response is inconsistent with a knee-jerk reflex: onset is delayed by roughly a second or more rather than immediate, both legs move, or the knee extension is accompanied by visible hip or ankle motion.'' \\
    3 points: ``Within the same continuous shot as the strike, the left lower leg extends forward at the knee with no perceptible delay after contact; the movement is confined to the knee joint, with the hip and ankle visibly uninvolved.'' \\
    4 points: ``All elements of 3-score are present AND the extension is a single brief kick rather than a slow push or an exaggerated fling, and the leg returns toward its resting position afterward without additional jerks, bounces, or oscillations.'' \\
    5 points: ``All elements of 4-score are present AND nothing suggests the subject is voluntarily assisting or resisting the movement (no visible muscle bracing, grimace, or hand moving toward the leg); the kick begins and ends cleanly with no tremor, sustained contraction, or rebound afterward, consistent with a brief involuntary contraction rather than a deliberate motion.'' \\
\} \\
\vspace{0.2em}

sc\_specification: \{ \\
    1 point: ``The subject, hammer, or leg jumps to a different position or shape between frames without continuous motion connecting them; the scene's lighting flips or changes abruptly partway through the shot; the left knee or lower leg bends or bulges in a way a real joint cannot.'' \\
    2 points: ``A visible cut or gap interrupts the hammer's path between its approach and the moment it reaches the knee; the direction of the light and shadows on the leg or hammer changes partway through the clip with no light source moving to explain it; part of the leg or hammer flickers in and out of view where nothing should be blocking it.'' \\
    3 points: ``The subject, hammer, and both legs stay in consistent relative positions throughout (the crossed-leg posture does not shift on its own); the hammer's path from raised position to contact is shown as one continuous motion; the light source and the resulting shadows on the leg and hammer keep the same direction and strength across the whole clip.'' \\
    4 points: ``All elements of 3-score are present AND the hammer's swing speeds up and slows down the way a real swung object would (not a constant-speed glide or a sudden jump into contact); if the leg extends, its arc follows a natural pivot at the knee rather than sliding in a straight line or bending at an impossible point.'' \\
    5 points: ``All elements of 4-score are present AND the camera framing stays fixed with no visible drift, wobble, or unexplained repositioning across the whole shot; the hammer never appears to pass through or behind the leg when it should be in front of it; the subject, hammer, and leg all appear to occupy consistent depth and scale relative to one another throughout.'' \\
\} \\
\vspace{0.2em}

\end{tcolorbox}
\caption{The evaluation specification of the knee-jerk reflex example, reproduced
verbatim from the released dataset.}
\label{fig: example-prompt_rubic}
\end{figure*}

%% file: appendix/experiment_results.tex
\clearpage
\section{Additional Experimental Results}

\subsection{Full-Benchmark Results}\label{app:full_results}
Table~\ref{tab: full-results} reports the complete per-dimension automatic evaluation results for open-source models on the full \ours benchmark. The corresponding overall averages are also included in the \textbf{Full Avg.} column of Table~\ref{tab: total-results}.
\input{tables/table_full}

\subsection{Error Analysis}
\label{sec: error analysis}
We conduct a detailed error analysis and classify the errors into three major categories as follows:
(1) \textbf{Poor Adherence to Instructions}: Our analysis reveals that even leading models struggle with instruction consistency in knowledge-intensive generation tasks.  
In particular, models demonstrate a lack of detail awareness, failing to generate fine-grained specifics mentioned in the prompt: in \autoref{fig:error_analysis_fig_1}, the model is unable to render the momentary push-button switch specified in the prompt, showing a bare fingertip touching the board instead.

(2) \textbf{Inaccurate Simulation of Scientific Principles:} A crucial weakness of current T2V models is their inability to generate content that respects scientific knowledge and the laws of the physical world. Models often fail to reason from preconditions and produce scientifically plausible outcomes. For example, as shown in \autoref{fig:error_analysis_fig_2}(a), a video meant to depict a knee-jerk reflex incorrectly shows the un-struck leg kicking forward.
Similarly, in \autoref{fig:error_analysis_fig_2}(b), the bag is squeezed directly beside the eye, yet the subject never blinks, demonstrating a broken understanding of the corneal reflex.
These errors indicate that models tend to prioritize visual aesthetics over physical and scientific realism. 

(3) \textbf{Deficiencies in Temporal Coherence and Visual Quality}: Beyond semantic and scientific inaccuracies, T2V models suffer from significant artifacts that break the illusion of realism. These issues primarily concern temporal consistency and overall visual quality. A major problem is the lack of object permanence and consistency; for instance, in \autoref{fig:error_analysis_fig_3}, the two balls released simultaneously on the two tracks merge into a single ball mid-motion. Similarly, models struggle to maintain a consistent visual style, as seen in \autoref{fig:error_analysis_fig_4}, where a realistic prism scene jarringly transitions to an animated style for the refracted light. In addition to these consistency failures, models also exhibit other common defects, such as poor image quality with coarse textures (\autoref{fig:error_analysis_fig_5}(a)) and weak causal links between events, like an LED switching on and off without any relation to the hand waving in front of the motion sensor (\autoref{fig:error_analysis_fig_5}(b)). Together, these flaws disrupt the visual flow and undermine the generated video's believability.

\input{appendix/Error_Analysis/error_analysis}

%% file: tables/table_full.tex
\begin{table}[H]
\centering
\footnotesize
\renewcommand{\arraystretch}{1.1}
\setlength{\tabcolsep}{4pt}
\resizebox{\textwidth}{!}{ \begin{tabular}{lcrrrrr}
\toprule[.1em]
\textbf{Models} & \textbf{Release} & \textbf{VT} & \textbf{SC} & \textbf{PG} & \textbf{SCC} & \textbf{Avg.} \\
\midrule
\multicolumn{7}{c}{\emph{\textbf{Open-source Models}}} \\
\midrule
MiniMax-H3~\cite{minimaxh3} & 2026-08 & 3.91 & 2.61 & \textbf{2.34} & \textbf{1.60} & \textbf{2.61} \\
Wan2.2-5B~\cite{DBLP:journals/corr/abs-2503-20314} & 2025-07 & 3.81 & \textbf{2.78} & 1.87 & 1.26 & \underline{2.43} \\
HunyuanVideo-1.5~\cite{DBLP:journals/corr/abs-2511-18870} & 2025-11 & 3.78 & \underline{2.78} & 1.83 & 1.32 & \underline{2.43} \\
Cosmos3-Nano~\cite{cosmos3} & 2026-06 & \textbf{4.02} & 2.70 & 1.72 & 1.27 & \underline{2.43} \\
LongCat-Video~\cite{DBLP:journals/corr/abs-2510-22200} & 2025-10 & 3.85 & 2.52 & 1.78 & 1.23 & 2.34 \\
LTX-2~\cite{DBLP:journals/corr/abs-2601-03233} & 2026-01 & \underline{4.01} & 2.49 & 1.53 & 1.20 & 2.31 \\
CogVideoX1.5-5B~\cite{DBLP:conf/iclr/YangTZ00XYHZFYZ25} & 2024-11 & 3.89 & 2.11 & \underline{1.88} & \underline{1.35} & 2.30 \\
LTX-2.3~\cite{DBLP:journals/corr/abs-2601-03233} & 2026-03 & 3.93 & 2.07 & 1.79 & 1.29 & 2.27 \\
\bottomrule[.1em]
\end{tabular}
}
\caption{Complete automatic evaluation results for open-source models on the full \ours benchmark (\nexample examples), sorted by the rightmost \textbf{Avg.} column. Metric definitions follow \Cref{tab: total-results}.}
\label{tab: full-results}
\end{table}

%% file: appendix/Error_Analysis/error_analysis.tex
\begin{figure}[htbp]
    \centering
    \includegraphics[width=\linewidth]{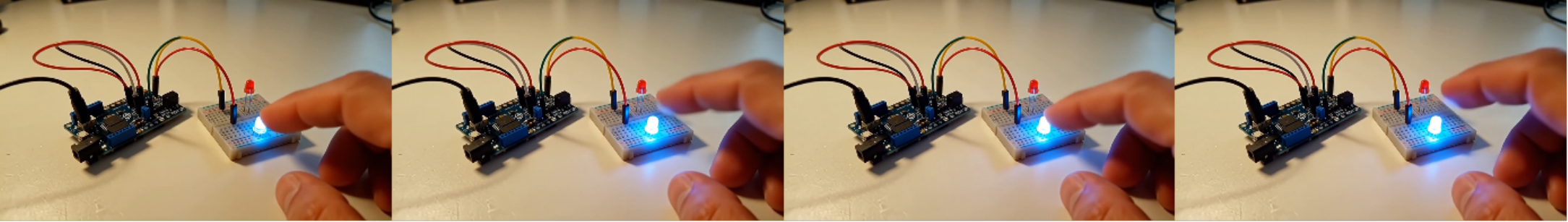}
    \vspace{0.6em}
    \small A desktop circuit demonstrates a momentary push-button switch connected to a power source and a single LED with a resistor, while a hand repeatedly presses and releases the button.
    \caption{Category (1), poor adherence to instructions: the video is inconsistent with the specified setup. The verbatim benchmark prompt is quoted below the example.}
    \label{fig:error_analysis_fig_1}
\end{figure}

\begin{figure}[htbp]
    \centering
    \includegraphics[width=\linewidth]{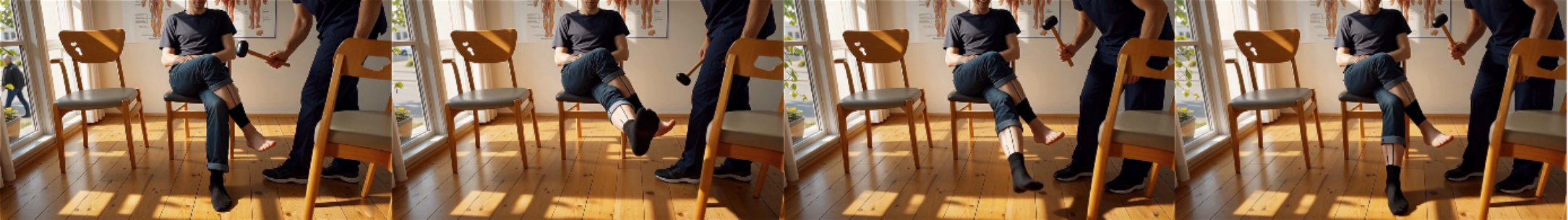}
    \vspace{1em}
    \small (a) A subject sits on the examination table with the left leg naturally placed over the right leg, while another person rapidly strikes the patellar tendon below the left knee with a rubber hammer.
    \includegraphics[width=\linewidth]{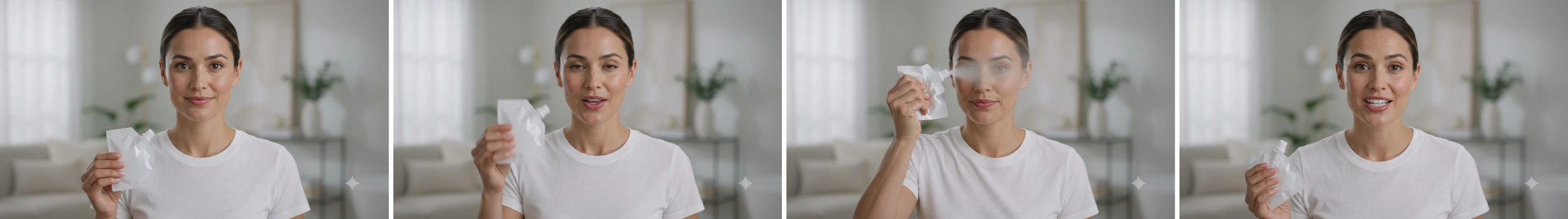}
    \vspace{1em}
    \small (b) A person is facing the camera, with one hand holding a bag and delivering a brief puff of air near one eye from the side.

    \vspace{0.6em}
    \caption{Category (2), inaccurate simulation of scientific principles: the required mechanism is not realized. The verbatim benchmark prompt is quoted below each example.}
    \label{fig:error_analysis_fig_2}
\end{figure}

\begin{figure}[htbp]
    \centering
    \includegraphics[width=\linewidth]{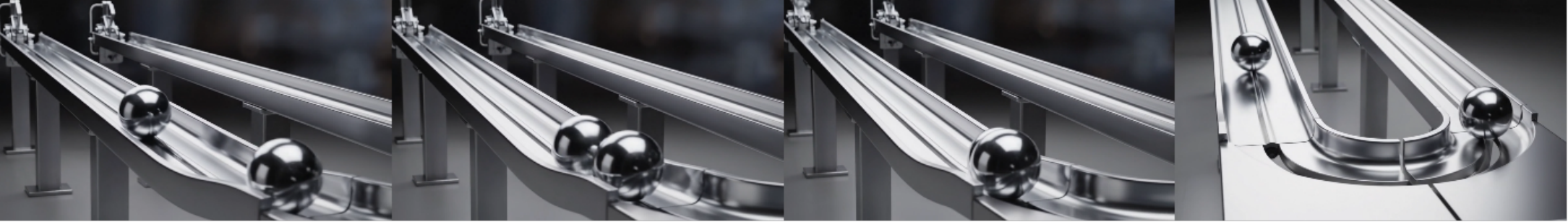}
    \vspace{0.6em}
    \small Within the same vertical plane, two fixed tracks are set up: one is a straight line, and the other is a cycloid-shaped curve. The starting and ending heights of both tracks are the same. Two identical small balls are released simultaneously from the starting point of the tracks, and the one that reaches the endpoint first is observed.

    \vspace{0.6em}
    \caption{Category (3), deficiencies in temporal coherence: objects lose permanence over time. The verbatim benchmark prompt is quoted below the example.}
    \label{fig:error_analysis_fig_3}
\end{figure}

\begin{figure}[htbp]
    \centering
    \includegraphics[width=\linewidth]{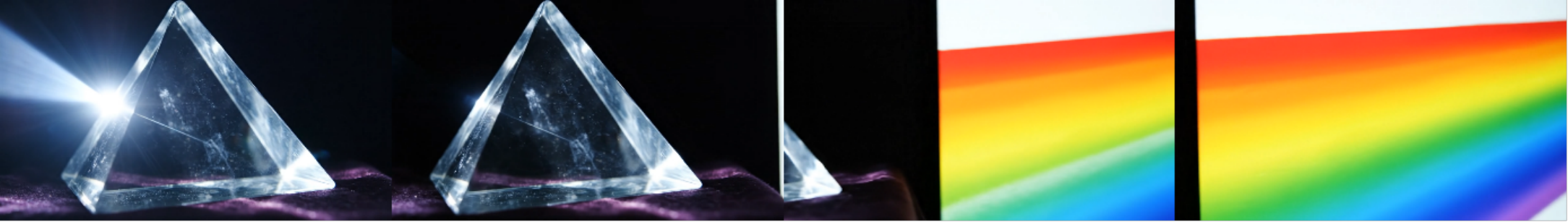}
    \vspace{0.6em}
    \small A triangular glass prism is placed in a beam of sunlight, allowing the light to pass through it. Observe what happens to the light as it exits the prism, noting the appearance, distribution, and order of any colors that result.

    \vspace{0.6em}
    \caption{Category (3), deficiencies in visual quality: the visual style shifts discontinuously. The verbatim benchmark prompt is quoted below the example.}
    \label{fig:error_analysis_fig_4}
\end{figure}

\begin{figure}[htbp]
     \centering
    \includegraphics[width=\linewidth]{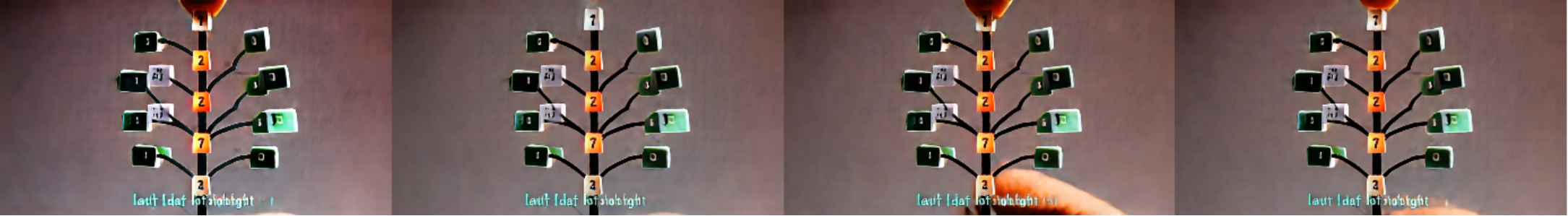}
    \vspace{1em}
    \small (a) Display a simple binary tree on a light background, with nodes labeled (1--7); highlight the nodes from top to bottom according to their level using bright colors (e.g., red) to demonstrate level order traversal. Briefly illuminate each node when visited, and add a mark for it (its numeral if legible digits can be rendered, otherwise a simple tally mark or dot) to the left-to-right output list at the bottom of the screen.
    \includegraphics[width=0.95\linewidth]{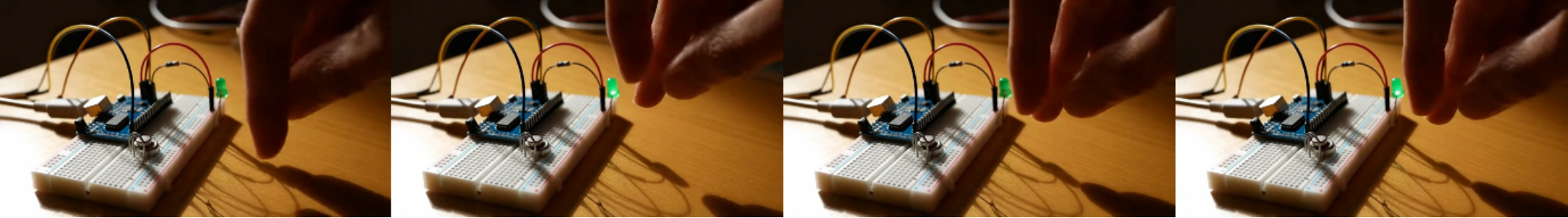}
    \vspace{1em}
    \small (b) On the workbench, a microcontroller is connected to a PIR motion sensor and an LED, and is powered on. A hand enters the frame and waves back and forth in front of the sensor for a few seconds before withdrawing.
    \caption{Further category (3) defects: coarse textures and weak causal links between events. The verbatim benchmark prompt is quoted below each example.}
    \label{fig:error_analysis_fig_5}
\end{figure}